\documentclass[letterpaper, 10 pt, conference]{ieeeconf}  

\IEEEoverridecommandlockouts                              

\usepackage{graphics} 
\usepackage{epsfig} 
\usepackage{mathptmx} 
\usepackage{times} 
\usepackage{amsmath} 
\usepackage{amssymb}  
\usepackage{tabularx}
\usepackage{algorithm,algpseudocode}
\usepackage{dblfloatfix}    
\usepackage{mathtools}
\usepackage{subcaption}
\usepackage{booktabs}
\usepackage{gensymb}
\usepackage{url}
\usepackage[noadjust]{cite}
\usepackage{xcolor}

\newcommand{\cca}[1]{{\color{black}{}#1}}
\newcommand{\ccn}[1]{{\color{black}{}#1}}

\newcommand{\shacamera}[1]{{\color{black}{}#1}}

\DeclareMathSymbol{@}{\mathord}{letters}{"3B}

\newcommand{\mypara}[1]{\vspace{2pt}\noindent\textbf{#1}}

\title{\Large \bf Relational Graph Learning for Crowd Navigation}

\author{Changan Chen\thanks{$^{*}$CC and SH contributed equally}\thanks{$^{\dagger}$work done as an undergrad at Simon Fraser University}$^{*\dagger1,2}$, Sha Hu$^{*2}$, Payam Nikdel$^{2}$, Greg Mori$^{2}$, Manolis Savva$^{2}$\\
$^{1}$UT Austin, $^{2}$Simon Fraser University
}

\begin{document}

\bstctlcite{IEEEexample:BSTcontrol}

\maketitle
\thispagestyle{empty}
\pagestyle{empty}

\pdfminorversion=4  
\begin{abstract}
We present a relational graph learning approach for robotic crowd navigation using model-based deep reinforcement learning that plans actions by looking into the future.
Our approach reasons about the relations between all agents based on their latent features and uses a Graph Convolutional Network to encode higher-order interactions in each agent's state representation, which is subsequently leveraged for state prediction and value estimation.
The ability to predict human motion allows us to perform multi-step lookahead planning, taking into account the temporal evolution of human crowds.
We evaluate our approach against a state-of-the-art baseline for crowd navigation and ablations of our model to demonstrate that navigation with our approach is more efficient, results in fewer collisions, and avoids failure cases involving oscillatory and freezing behaviors.
\end{abstract}

\section{INTRODUCTION} \label{sec:intro}

Inferring the underlying relations between components of complex dynamic systems can inform decision making for autonomous agents.
One natural system with complex dynamics is crowd navigation (i.e., navigation in the presence of multiple humans). 
The crowd navigation task is challenging as the agent must predict and plan relative to likely human motions so as to avoid collisions and remain at safe and socially appropriate distances from people.
Some prior work predicts human trajectories using hand-crafted social interaction models \cite{helbing_social_1995} or by modeling the temporal behavior of humans \cite{alahi_social_2016}. Although these methods can estimate human trajectories, they do not use the prediction to inform the navigation policy.
Other recent works \cite{chen_decentralized_2016,everett_motion_2018, chen_crowd-robot_2018} use deep reinforcement learning (RL) to learn a socially compliant policy. 
These policies either do not leverage the human interactions or approximate it with heuristics. They also simplify the human motion prediction problem with unrealistic assumptions such as linear human motion, and typically consider only the current state of humans instead of incorporating predicted human motion to inform the policy.

\begin{figure}
\centering
\includegraphics[width=0.9\linewidth]{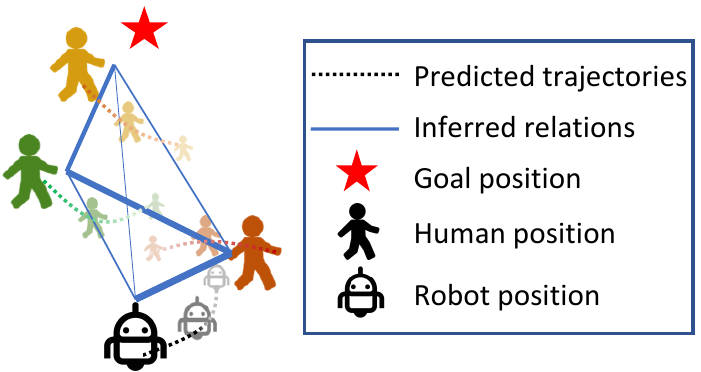}
\caption{Illustration of our relational graph learning approach (relation graph in blue). We infer interactions among the robot and humans and predict trajectories (line thickness indicates the strength of interaction; large/medium/small figures show current positions, the predicted position in next step, and the predicted position in two steps). By inferring a strong relation between the robot and the red human as well as the green and red human, and planning conditional on predicted human trajectories, we can find a safe path towards the goal.}
\label{fig:pull-figure}
\end{figure}

More broadly, interacting systems have been studied extensively in recent work \cite{deng_structure_2015, kipf_neural_2018,battaglia_interaction_2016} and 
Graph Neural Networks (GNNs) are one of the most powerful tools for modeling objects and their relations (interactions).
A variant of GNNs is Graph Convolutional Networks (GCNs) \cite{kipf_semi-supervised_2016} where relations between nodes are defined as an adjacency matrix.
Whereas in GCNs the relations between all nodes are given, Wang et al.~\cite{wang_non-local_2017} and Grover et al.~\cite{grover_learning_2018} propose to learn the relations between objects and use learned attention to compute new features.
Inspired by this work on relational reasoning and GNN models, we propose a relational graph learning model for crowd navigation that reasons about relations between agents (robot and humans) and then use a GCN to compute interactions between the agents.
With the predicted interaction features of both the robot and humans, our approach jointly plans efficient robot navigation and predicts the motion of present humans.
Figure \ref{fig:pull-figure} illustrates how interaction reasoning between all agents and explicit prediction can yield a farsighted navigation policy.

Planning and prediction of multi-agent states is fundamental in many problems, including the general case of decision making in an environment with $N$ non-cooperative agents.
In this paper, we address this problem with a relational graph learning approach for model-based RL, which predicts future human motions and plans for crowd navigation simultaneously.
We show that our model outperforms baselines from prior work and carry out ablation studies to measure the planning budget on navigation performance.
We also qualitatively demonstrate that our approach mitigates undesirable robot behaviors in challenging scenarios.
The code of our approach is publicly available at \url{https://github.com/ChanganVR/RelationalGraphLearning}.

\raggedbottom
\section{Related Work}
\mypara{Crowd navigation.}
Mobile robot navigation in crowds is challenging due to the complex ways in which human intent and social interactions determine human motions.
Prior work has used rule-based algorithms to characterize the interactions between agents.
Helbing et al.~\cite{helbing_social_1995} proposed to model the interactions as "social forces".
RVO~\cite{berg_reciprocal_2008} and ORCA~\cite{van_den_berg_reciprocal_2011} solve for collision avoidance under reciprocal assumptions.
Interacting Gaussian Process (IGP) \cite{trautman_unfreezing_2010, trautman_robot_2013, trautman_sparse_2017}  model the trajectory of each agent as an individual Gaussian Process and propose an interaction potential term to couple the individual GP for interaction.
Rule-based methods rely heavily on hand-crafting the model for navigation.
Social-LSTM~\cite{alahi_social_2016} used an LSTM to predict human trajectories directly from large scale datasets.
However, for crowd navigation, forecasting models do not directly yield an action policy.
\cca{Prediction-based models~\cite{aoude_probabilistically_2013, bennewitz_learning_2005} perform motion prediction and planning sequentially but face the freezing robot problem in complex environments.}
Recently, Chen et al.~\cite{chen_decentralized_2016, chen_socially_2017} propose to use deep RL for crowd navigation by learning a value network encoding the state values.

LM-SARL~\cite{chen_crowd-robot_2018} improved on previous work by learning the robot state representation with attentive pooling over pairwise interaction features.
However, this model is limited by partial modeling of crowd interactions, due to significant simplifying assumptions for the underlying state transition without explicitly modeling human motions.
Most recently, LeTS-Drive~\cite{cai_lets-drive:_2019} used online belief-tree search to learn a value and policy function for autonomous driving in a crowded space.
Although this approach models intentions and interactions between the vehicle and humans, the interaction is coarse-grained, utilizing Convolutional Neural Networks (CNNs) to process stacked frames of the environment, the history and intended path of the vehicle.
In contrast, we include pairwise interactions among all agents, which coupled with our graph learning, explicitly captures relations between agents and models higher-order interactions.

\mypara{Relational reasoning.}
Relational reasoning~\cite{battaglia_relational_2018, vaswani_attention_2017} aims to capture relationships between entities such as decision-making agents~\cite{zambaldi_relational_2018, agarwal_learning_2019}, image pixels~\cite{wang_non-local_2017}, words\cite{lin_structured_2017}, or humans and objects \cite{chao_hico:_2015, gkioxari_detecting_2017}.
The relations and entities are typically represented as a connected graph \cite{xu_scene_2017, velickovic_graph_2017}, and standard tools for graph-based learning such as Graph Neural Nets(GNNs)~\cite{hamilton_representation_2017, xu_how_2019, kipf_semi-supervised_2016, hamilton_inductive_2017} are applied.
GNNs are a class of neural networks that learn functions on graphs, representing objects as nodes and relations as edges.
Knowledge about object dynamics can be encoded in the GNN node update function, and interaction dynamics can be encoded in the edge update function.
Most current GNNs operate on interacting systems with known and fixed graph inputs, such as human skeletons or particles connected by springs~\cite{battaglia_interaction_2016, kipf_semi-supervised_2016}.
However, many interacting systems have unknown relations. For example, humans in sports~\cite{ibrahim2018hierarchical} and crowds~\cite{vemula_social_2017}.
Thus, it is important to infer relations among the entities and learn based on the inferred relations. 
In crowd navigation, the temporal dynamics of these relations are useful for planning safe and efficient crowd navigation (e.g., understanding the joint motion of a pair of friends as they cross the robot's path while walking close to each other, in contrast to the motion of a pair of strangers walking in opposite directions and passing each other).
Inspired by Kipf et al.~\cite{kipf_neural_2018}, who estimate the graph connectivity map from trajectories using an auto-encoder architecture, our model dynamically infers the crowd-robot relation graph at each time step and learns the state representation for each agent based on this graph.
\cca{
Recent work~\cite{chen_robot_2019} proposed to use graph convolutional networks in navigation and used human gaze data to train the network. The use of human gaze data helps the network to learn more human-like attention but it is also limited to the robot's attention. In our proposed method, the GCN not only captures the attention of the robot but also inter-human attention, which is subsequently leveraged by a human motion prediction model.
}

\mypara{MPC, MCTS and model-based RL.}
Model predictive control (MPC) is a family of control algorithms that leverage models of the system to predict state changes and plan control accordingly.
Traditional MPC~\cite{wieber_trajectory_2006, mayne_robust_2005} usually assumes access to a known environment model, which is frequently unrealistic.
Monte-Carlo Tree Search (MCTS) has been used for decision-time planning by estimating action values based on many simulated trajectories in complex search problems such as the game of Go~\cite{silver_mastering_2016}.
More recent model-based RL methods first acquire a predictive model of the world, and then use that model to make decisions.
Finn et al.~\cite{finn_deep_2016} learned a state transition model by predicting future frames to achieve goal-directed robotic manipulation.
VPN~\cite{oh_value_2017} and Predictron~\cite{silver_predictron:_nodate} learned to predict future abstract states that are informative for planning in Atari Games.
In contrast, our model's predictive relational graph takes a set of raw human states (e.g., positions, velocities) as input and predicts multiple interacting human trajectories.
To our knowledge, we are the first to integrate relational learning with a model-based RL algorithm for crowd navigation.

\begin{figure}[t]
    \centering
    \includegraphics[width=1\linewidth]{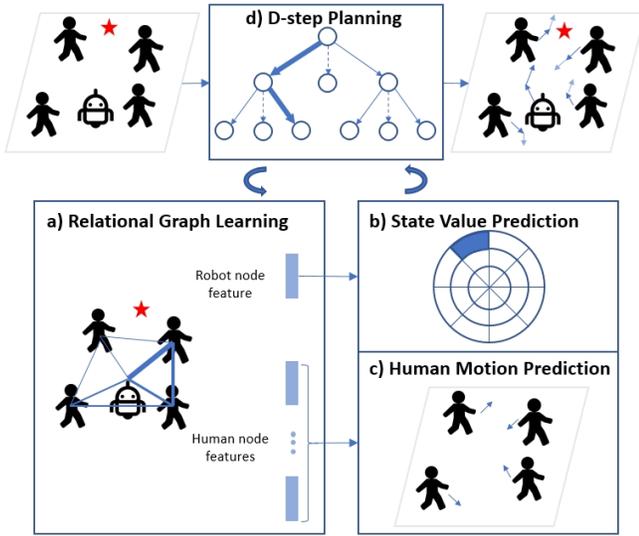}
    \caption{Given the crowd state as input, our approach uses $d$-step planning to discover the best action sequence for safe and efficient crowd navigation:
    a) a relational graph learning model reasons about relations between agents and encodes local interactions. b) a value network predicts the value of the robot state representation. c) a motion network predicts future human states. d) with the learned value estimation and motion prediction models, our approach rolls out d steps into the future and searches for the best action sequence.}
    \label{fig:model}
\end{figure}


\section{Approach}
\label{sec:approach}

We first describe how we formulate the crowd navigation problem with deep RL, then introduce our relational graph learning model for modeling interactions in the crowd. 
\cca{In addition, we show how this model can be augmented by a planning algorithm (simplified MCTS) at both training and test time.}
Figure \ref{fig:model} shows an overview of our approach.

\subsection{Deep Reinforcement Learning for Crowd Navigation}
In this work, we address the crowd navigation task where the robot navigates through a crowd of $N$ humans to a goal position as efficiently and safely as possible.
This task is formulated as a sequential decision making problem in recent works \cite{chen_decentralized_2016, everett_motion_2018, chen_crowd-robot_2018}.
Each agent (either human or the robot) observes others' observable state, including position $\mathbf{p}=[p_x,p_y]$, velocity $\mathbf{v}=[v_x,v_y]$ and radius $r$ (an abstract measure of size).
Each agent also maintains an internal state, such as a goal position $\mathbf{p_g}$ and preferred speed $v_{pref}$.
We assume actions are instantaneous and the agent always arrives at the target position in the next time step.
We use $s_0^t$ and $s_i^t$ to denote the robot state and the observed state of human $i$ at time $t$, respectively.
The robot input state is defined as $\mathbf{S}^t = \{s_0^t, s_1^t, ..., s_N^t\}$.
The optimal policy mapping state $\mathbf{S}^t$ to action $\mathbf{a}^t$ at time $t$, $\pi^* : \mathbf{S}^t \mapsto \mathbf{a}^t$, is to maximize the expected return:

\begin{equation}\label{eq:rl}
\begin{aligned}
\pi^{*}(\mathbf{S}^t) = & \underset{\mathbf{a}^t}{\text{argmax}} ~ R(\mathbf{S}^t,\mathbf{a}^t) + \\
& \gamma^{\Delta t \cdot v_{pref}} \int_{\mathbf{S}^{t+\Delta t}} P(\mathbf{S}^t,\mathbf{a}^t,\mathbf{S}^{t+\Delta t})  V^*(\mathbf{S}^{t+\Delta t}) d\mathbf{S}^{t+\Delta t} \\
V^*(\mathbf{S}^t) = & \sum_{k=t}^T \gamma^{k \cdot v_{pref}} R^{k}(\mathbf{S}^{k},\pi^*(\mathbf{S}^{k})), \\
\end{aligned}
\end{equation}

where $R(\mathbf{S}^t,\mathbf{a}^t)$ is the reward received at time $t$, $\gamma \in (0,1)$ is the discount factor, $V^*$ is the optimal value function, $P(\mathbf{S}^t,\mathbf{a}^t,\mathbf{S}^{t+\Delta t}) $ is the transition probability from time $t$ to time $t+\Delta t$. And the preferred velocity $v_{pref}$ is used as a normalization term in the discount factor for numerical reasons~\cite{chen_decentralized_2016}.

\cca{
In the above equation, $P(\mathbf{S}^t,\mathbf{a}^t,\mathbf{S}^{t+\Delta t}) $ represents the system dynamics and the knowledge of the world (e.g., how state changes depend on agent actions) and is usually unknown to the agent. Some recent work~\cite{chen_decentralized_2016, chen_crowd-robot_2018} assumes the state transition function to be known during training time and models it with simple linear models at test time. This assumption strongly reduces the complexity of the problem since the agent can basically solve the navigation problem by searching the next state space. In this work, we remove the assumption of knowing the state transition, and instead use a model-based approach to learn to predict human motions.
}

We follow the formulation of the reward function defined in Chen et al.~\cite{chen_crowd-robot_2018}, which awards accomplishing the task while penalizing collisions or uncomfortable distances.

This problem statement can be applied to a set of more general tasks where there are $N$ non-cooperative agents and the decision-maker only receives their observable states but does not know about their intents or hidden policies.

\subsection{Model Predictive Relational Graph Learning}
The interactions (i.e., spatial relations) between humans are important for robot navigation and for predicting future human motion.
Previous work does not learn such interaction features for the robot and humans simultaneously.
Here, we model the crowd as a graph, reason about the relations between all agents and use a GCN to compute the robot and human state representations.
Using the graph model as a building block, we further construct two other modules: a \emph{value estimation module} $f_V()$ which estimates the value of the current state and a \emph{state prediction module} $f_P()$ which predicts the state at the next time step.

\mypara{Relational graph learning.}
The key challenge in crowd navigation is to learn a good representation of the crowd encoding interactions among all agents.
Chen et al.~\cite{chen_crowd-robot_2018} show that attentive crowd aggregation improves both interpretation and performance by modeling one-way human-robot interactions.
This motivates us to model the crowd and the robot as a directed graph $G^t=(V^t,E^t)$ where $|V|=N+1$.
The edge $e_{i,j} \in E$ indicates how much attention agent $i$ pays to agent $j$ or the importance of agent $j$ to agent $i$.
This pairwise relation is not known a priori, so it is inferred with a pairwise similarity function (relation inference).
After the relations between all agents are inferred, a graph neural network propagates information from node to node and computes the state representations for all agents (interaction modeling) as shown in Figure \ref{fig:model} a).

LM-SARL~\cite{chen_crowd-robot_2018} can be viewed as an approximation of our Relational Graph Learning (RGL) formulation, in that it learns robot interaction features with respect to all humans and uses attentive pooling to aggregate the interaction features.
Our RGL formulation not only learns the attention of the robot to humans but also from humans to humans.
Apart from learning the robot state representation, RGL also learns the state representation for other agents simultaneously and propagates these features to the robot node by message passing.
In this way, RGL also models higher-order interactions.
For example, $l=2$ fully models human-human interaction rather than using a local map (as in LM-SARL) to approximate the local interactions of humans.
This approach is also favorable compared to LeTS-Drive~\cite{cai_lets-drive:_2019} in that LeTS-Drive doesn't reason about pairwise state interactions or model human-human interactions. 
\subsubsection{Relation Inference}
the initial values of vertices $V$ are the state values for the agents:  $S^t = \left \{ s^{t}_{i} \right \}_{i=0...N+1}$.
Since robot and human states have different dimensions, we use two multilayer perceptrons (MLPs) $f_r()$ and $f_h()$ to embed them into a latent space, resulting in a matrix $X$, where the first row is the latent state of the robot and the remaining rows are the latent states of humans.
Given the feature matrix $X$, a relation matrix is computed using a pairwise similarity function. Following Wang et al.~\cite{wang_non-local_2017}, we use an embedded Gaussian as the similarity function.
The pairwise form is given by $f(x_i, x_j) = e^{\theta(x_i)^T \phi(x_j)}$ and the matrix form is given by $A = \text{softmax}(XW_aX^T)$ where $x_i = X[i, :]$, $\theta(x_i)=W_\theta x_i$, $\phi(x_i)=W_\phi x_i$ and $W_a = W_\theta W_\phi^T$. A learned relation is illustrated in Figure~\ref{fig:pull-figure} where the thickness of the line indicates the strength of pairwise interactions.

\subsubsection{Interaction Modeling}
with the feature matrix $X$ and relation matrix $A$, we use a GCN to compute the pairwise interaction features.
The message passing rule is defined by $H^{(l+1)} = \sigma(A H^{(l)}W^{(l)}) + H^{(l)}$ where $W^{(l)}$ is a layer-specific trainable weight matrix, $H^{(l)}$ is the node-level feature of layer $l$, and $\sigma$ is an activation function. The feature of node $i$ at level $l+1$ aggregates its neighbor node features at level $l$ weighted by the relations stored in matrix $A$.

Let $H^{(0)} = X$ and after $L$ layer propagations, we have state representation matrix $Z^t = H^{(L)}$ for $S^t$, and $Z^t[i, :]$ is the state representation for agent $i$ encoding its local interactions.

\mypara{Relational value estimation.}
Our value estimation module $f_V$ consists of two models: a relational graph model to infer the robot state representation $Z^t[0, :]$ and a subsequent value network to predict the value of the state $v = f_{v}(Z^t[0, :])$.
The value network $f_v$ is an MLP and predicts values over a discretized action space by looking ahead as shown in Figure \ref{fig:model} b).

\mypara{State prediction.}
We assume the action of the robot can be perfectly executed and the next robot state $\hat{s_0}^{t+1}$ can be computed directly from the current state $s_0^{t}$ and action $a^t$.
Our state prediction module models interactions between human states to predict future human states, as shown in Figure \ref{fig:model} c).
In the action selection phase, previous works~\cite{chen_decentralized_2016, chen_crowd-robot_2018} rely on either querying the ground truth or off-the-shelf algorithms to approximate the next human states, which does not benefit from end-to-end learning.
Our state prediction module $f_P()$ consists of two models: first a relational graph model predicts relations between all agents and their $l^{th}$ layer interaction features, then a successor motion prediction model uses the human state representations to predict their next state $\hat{s_i}^{t+1} = f_{m}(Z^t[i, :])$, where $\hat{s_i}^{t+1}$ is the predicted state for human $i$ with $1\leq i \leq n $. The motion prediction network $f_m$ is modelled by an MLP.
In addition, to simplify the prediction task, we use a hand-crafted function to estimate the reward based on the prediction of human states, denoted by $\hat{R}(S^t, a^t, \hat{S}^{t+1})$.
\cca{By jointly learning state values and predicting human motions, the policy learning benefits from encoding human motions and better addresses the freezing robot problem in \cite{aoude_probabilistically_2013, bennewitz_learning_2005}.}

\subsection{Relational Forward Planning and Learning}
In this section, we demonstrate how relational graph learning can be augmented with a simplified MCTS approach to provide bootstrapped targets in the learning phase and a far-sighted policy at decision time.
\cca{Imperfect learned value functions can lead to suboptimal actions due to local minima.
To leverage the prediction ability of our model, we further simulate $d$-steps into the future to provide a better estimate of the state values.
}
Furthermore, by controlling the depth $d$ of this simulation, we can trade off computation for better performance.
Our approach is visualized in the tree search diagram shown in Figure \ref{fig:model} d).

We follow the $d$-step planning method proposed in Oh et al.~\cite{oh_value_2017}, performing rollouts using the crowd state prediction and value estimation up to $d$ steps in the future, and select the action with the maximum $d$-step predicted return, defined as follows:
\vspace{-0.1cm}
\begin{equation} \label{eq:d-step return}
\begin{aligned}
    V^d(S^t) = 
\begin{cases}
    \vspace{0.15cm} f_V(S^t),                                      & \text{if } d = 1\\ 
    \frac{1}{d}V^1(S^t) + \frac{d-1}{d} \max_{a^t}( &  \\
    \;\; \hat{R}(S^t, a^t, \hat{S}^{t+1}) + \gamma V^{d-1}(\hat{S}^{t+1}))  & \text{otherwise}
\end{cases}
\end{aligned}
\end{equation}
where $\hat{S}^{t+1}=f_P(S^t, a^t)$.

With $d$-step planning, computation increases exponentially with search depth and width.
Due to our large action space, we use action space clipping to reduce the computational cost.
Intuitively, the value function estimates the quality of entering a state. 
Using this estimate, we recursively search the top-$w$ next states with one-step lookahead.
Compared to only considering the top action, $d$-step planning provides a better state value estimate when the agent encounters unseen states.
Figure \ref{fig:model} d) shows one toy example of tree search with depth $d=2$ and clipping width $w=2$.

\subsection{Joint Value Estimation and State Prediction Learning}
Pseudocode for the joint state prediction and value estimation learning scheme is in Algorithm \ref{alg:joint-model-v-learning}.
Similar to the training scheme in Chen et al.~\cite{chen_crowd-robot_2018}, we first use imitation learning with collected experience from a demonstrator ORCA~\cite{van_den_berg_reciprocal_2011} policy to initialize the model, and then use RL to refine the policy.
\ccn{Imitation learning is important for policy initialization due to the sparse rewards in navigation, without which RL can't converge.}
We train $f_V$ using RL and $f_P$ with supervised learning. To stabilize training, we also use separate the graph models for these two functions.
By integrating the planning into the learning phase, our learning algorithm is able to uses bootstrapped state value estimations as the targets to learn a more accurate value function than Chen et al.~\cite{chen_crowd-robot_2018}.

\begin{figure}
\begin{algorithm} [H] 
\small
\captionsetup{font=small}
\caption{Learning for $f_P$ and $f_V$}
\begin{algorithmic}[1] 
\State Initialize $f_{P}, f_{V}$ with demonstration $\mathsf{D}$ \label{line:init-v-p}
\State Initialize target value network $\hat{f_{V}} \leftarrow f_{V}$
\State Initialize experience replay memory $\mathsf{E} \leftarrow \mathsf{D}$ \label{line:init-memory}
\For{episode = 1, M} \label{line:rl-start}
\State Initialize random sequence $S^{0}$
\Repeat
\State $a_t\leftarrow \text{argmax}_{a_t \in A} \hat{R}(S^t, a^t, \hat{S}^{t+1}) + \gamma^{\Delta t \cdot v_{pref}} V^d(\hat{S}^{t+1})$ \label{line:action_selection} \hspace{2cm} where $\hat{S}^{t+1} =f_{P}(S^t, a^t)$

\State Execute $a_t$ and obtain $r_t$ and $S^{t + \Delta t}$
\State Store tuple ($S^{t}, a^t, r^t, S^{t+\Delta t})$ in $\mathsf{E}$ 

\State Sample random minibatch tuples from $\mathsf{E}$
\State Set target for value network: $y_i = r_i + \gamma^{\Delta t \cdot v_{pref}}  \hat{V}^{d}({S}_{i+1})$
\State Update $f_{V}$ by minimizing $L_1 = ||f_{V}(S_{i}) - y_i||$  \label{line:value}
\State Set target for prediction network: $S_{i + 1}$ \label{line:train_prediction1}
\State Update $f_{P}$ by minimizing $L_2 = ||f_P(S_{i}, a_{i}) - S_{i+1}||$ \label{line:train_prediction2}
\Until terminal state ${s}_t$ or $ t \ge t_{max}$
\State Update target value network $\hat{f_{V}} \leftarrow f_{V}$
\EndFor \label{line:rl-end}
\State \textbf{return} $f_{P}, f_{V}$
\end{algorithmic}
\label{alg:joint-model-v-learning}
\end{algorithm}
\vspace{-1em}
\end{figure}


\begin{table*}
\centering
\begin{tabular}{@{}lccccc@{}}
\toprule
Method                                         & Success $\uparrow$       & Collision $\downarrow$        & Extra Time $\downarrow$ & Avg. Return $\uparrow$   & Max Diff. $\downarrow$                   \\ \midrule
ORCA \cite{berg_reciprocal_2008}                & $0.43 \pm 0.00$ & $0.57 \pm 0.00$ & $2.93 \pm 0.00$ & $0.081 \pm 0.000$ &  $0.604 \pm 0.000$       \\
LM-SARL-Linear \cite{chen_crowd-robot_2018}     & $0.90 \pm 0.02$ & $0.09 \pm 0.02$ & $3.15 \pm 0.24$ & $0.506 \pm 0.018$ &  $0.179 \pm 0.018$       \\
RGL-Linear (Ours)                               & $0.92 \pm 0.02$ & $0.04 \pm 0.03$ & $2.35 \pm 0.13$ & $0.541 \pm 0.014$ &  $0.144 \pm 0.014$   \\
MP-RGL-Onestep    (Ours)                        & $0.93 \pm 0.02$ & $0.03 \pm 0.02$ & $2.15 \pm 0.13$ & $0.551 \pm 0.025$ &  $0.134 \pm 0.025$   \\
MP-RGL-Multistep (Ours)                         & $\mathbf{0.96} \pm 0.02$ & $\mathbf{0.02} \pm 0.01$ & $\mathbf{1.86} \pm 0.07$ & $\mathbf{0.591} \pm 0.009$ & $\mathbf{0.094} \pm 0.009$      \\
\bottomrule
\end{tabular}
\caption{Quantitative results in circle crossing with five humans. The metrics are defined as follows: ``Success'':~the rate of robot reaching its goal without a collision; ``Collision'':~the rate of robot colliding with humans; ``Extra Time'':~extra navigation time to reach goal in seconds; ``Avg. Return'':~returns averaged over steps across episodes and all test cases.  ``Max Diff.'':~the difference between actual average returns and the upper bound of average returns. $\pm$ indicates standard deviation measured using five independently seeded training runs.}
\label{tab:circle_5humans}
\vspace{-2em}
\end{table*}

\section{Experimental Results}
\label{sec:result}

\label{icra_added}

\mypara{Implementation details.}
The hidden units of $f_r(\cdot)$, $f_h(\cdot)$, $f_v(\cdot)$, $f_m(\cdot)$ have dimensions ($64$, $32$), ($64$, $32$), ($150$, $100$, $100$), ($64$, $32$) and the output dimension of $W^{(l)}_{a}$ is $32$ for all $l=1...L$.
For fair comparison with the baseline method, we use the value network for the baseline as reported in Chen et al.~\cite{chen_crowd-robot_2018}.
All the parameters are trained using Adam~\cite{kingma_adam:_2014}, and the learning rate is $0.001$.
The discount factor $\gamma$ is set to be $0.9$.
The exploration rate of the $\epsilon$-greedy policy decays from $0.5$ to $0.1$ linearly in the first $5k$ episodes.
We assume holonomic kinematics for the robot, i.e. it can move in any direction. The action space consists of $80$ discrete actions: $5$ speeds exponentially spaced between (0, $v_{pref}$] and $16$ headings evenly spaced between [0, 2$\pi$).

\mypara{Simulation setup.}
We use the CrowdNav\footnote{https://github.com/vita-epfl/CrowdNav} simulation environment.
In this simulation, humans are controlled by ORCA~\cite{van_den_berg_reciprocal_2011}, the parameters of which are sampled from a Gaussian distribution to introduce behavioral diversity.
We use circle crossing scenarios for our experiments.
Circle crossing scenarios have $N=5$ humans randomly positioned on a circle of radius $4m$ with random perturbation added to their $x$,$y$ coordinates. 
The maximum speed for the agent, $v_{pref}$ is 1 $m/s$, and the goal is fixed so that it can be reached in a minimum time of $8$ seconds.
To fully evaluate the effectiveness of the proposed model, we look into the simulation setting where the robot is invisible to humans.
As a result, the simulated humans react only to humans but not to the robot.
This setting is a clean testbed for validating the model's ability to reason about human-robot and human-human interaction without affecting human behaviors.
All models are evaluated with $500$ random test cases.

\mypara{Quantitative Evaluation.}
As expected, the ORCA method fails badly in the invisible setting due to the violation of the reciprocal assumption.
The state-of-the-art method LM-SARL~\cite{chen_crowd-robot_2018} assumes that the next ground truth state is given during training.
To demonstrate the effectiveness of our relational graph learning model, we use a linear motion model (agents keep velocities as in the last state) in LM-SARL as well as serving as our state prediction model so that the comparison is purely in robot state representation learning. 
These two models are indicated by LM-SARL-Linear and RGL-Linear, respectively.
We refer to our full model as MP-RGL-Multistep and the model with one-step lookahead as MP-RGL-Onestep.
For all RGL-based models, we use a two-layer GCN. For MP-RGL-Multistep, we let $d=2$ and $w=2$.
We do not compare with LeTS-Drive \cite{cai_lets-drive:_2019} as it focuses on components other than interaction modeling and assumes different inputs. 
Table~\ref{tab:circle_5humans} reports the rates of success, collision, the average extra navigation time (defined as extra navigation time beyond the minimum possible 8 seconds) as well as average return, which is the cumulative discounted rewards averaged over all steps in the 500 test cases. 
To provide a clearer interpretation of the average return metric, we add one more metric, that is the difference between the average return and its upper bound.
The upper bound only exists in imaginary cases where there are no other humans and the agent can head straight to the goal all the time. 

As expected, the performance of LM-SARL-Linear drops compared to the one reported in~\cite{chen_crowd-robot_2018} after replacing the unrealistic assumption of access to the next ground truth state with a linear motion model.
The first comparison between LM-SARL-Linear and RGL-Linear reflects the improvement brought by modeling the crowd interaction with graph neural networks.
Then the comparison between RGL-Linear and MP-RGL-Onestep shows that using a state prediction model leads to a better estimate of human motions and thus yields a better navigation policy.
Finally, the best model in the table is our MP-RGL-Multistep.
Comparing it to the one-step version, we see that $d$-step planning improves the success rate, reduces extra time to reach the goal and increases the average return.
Even the best-performing model is not collision-free.
The robot is invisible to humans and the policy does not have access to the next ground truth state, making some collisions unavoidable (e.g., when the robot is surrounded by humans converging towards it).

\mypara{Effect of planning budget.}
In our $d$-step planning, both the tree depth $d$ and action space clipping width $w$ influence the computation time and planning performance.
With larger $d$, our approach is able to look further into the future and plan further ahead.
With larger $w$, we consider more actions at each depth and can predict a wider range of possible outcomes, potentially reaching a better path.
We study how the planning budget in $d$-step planning influences performance.
We tested the MP-RGL-Onestep model in Table~\ref{tab:circle_5humans} with various test-time planning budgets.
By simply setting $d=2, w=2$, the extra time decreases from $2.15$ to $2.06$ and the average return improves from $0.551$ to $0.572$.
With larger $d, w$, the performance is also further improved.
From these two experiments, we conclude that our $d$-step planning leads to better performance both at learning and decision-making.

\begin{figure}
  \centering
  \begin{subfigure}[b]{0.45\linewidth}
    \centering
    \includegraphics[width=0.9\textwidth]{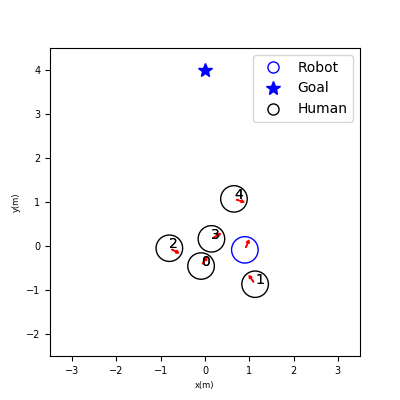}
    \caption{Scenario 1.}
    \label{fig:fig3a}
  \end{subfigure}
  \hfill
  \begin{subfigure}[b]{0.45\linewidth}
    \centering
    \includegraphics[width=0.9\textwidth]{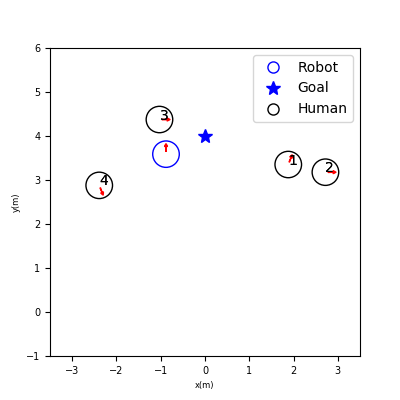}
    \caption{Scenario 2.}
    \label{fig:fig3d}
  \end{subfigure}
  \hfill
  \begin{subfigure}[b]{0.45\linewidth}
    \centering
    \includegraphics[width=\textwidth]{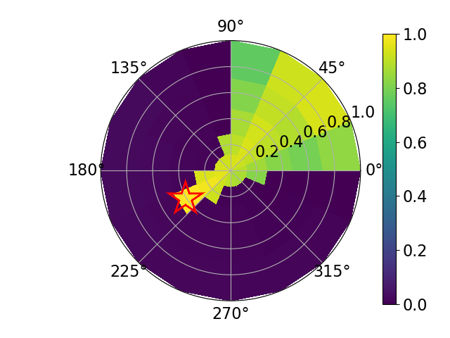}
    \caption{Value heatmap for $d=1$.}
    \label{fig:fig3b}
  \end{subfigure}
  \hfill
    \begin{subfigure}[b]{0.45\linewidth}
  \centering
    \includegraphics[width=\textwidth]{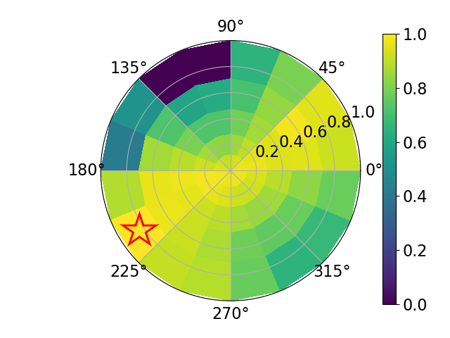}
    \caption{Value heatmap for $d=1$.}
    \label{fig:3e}
  \end{subfigure}
  \hfill
  \begin{subfigure}[b]{0.45\linewidth}
  \centering
    \includegraphics[width=\textwidth]{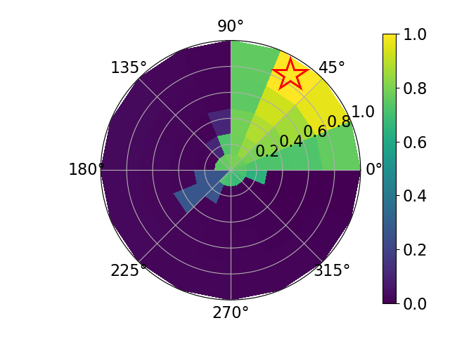}
    \caption{Value heatmap for $d=2$.}
    \label{fig:3c}
  \end{subfigure}
  \hfill
  \begin{subfigure}[b]{0.45\linewidth}
  \centering
    \includegraphics[width=\textwidth]{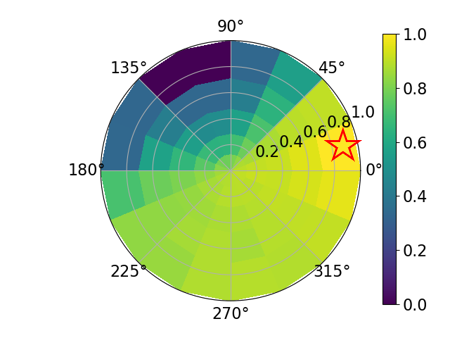}
    \caption{Value heatmap for $d=2$.}
    \label{fig:3f}
  \end{subfigure}
\caption{Value estimates for two scenarios using different planning depths $d$. $(\textrm{a}), (\textrm{c}), (\textrm{e})$ for Scenario 1 and $(\textrm{b}), (\textrm{d}), (\textrm{f})$ for Scenario 2.
The top row shows the top-down view of the crowd, and the two heatmaps below visualize the estimated values over the action space of the robot. The red star in the value map indicates the action with the highest value.}
\label{fig:vf}
\end{figure}

\mypara{Investigation of \cca{computation time}.}
The GNN-based state feature extractor is computationally efficient compared to attention-based crowd navigation work~\cite{chen_crowd-robot_2018} \ccn{due to sequential pairwise interaction feature extraction}.
The GNN computation amounts to a matrix multiplication, with negligible change when the number of agents N is relatively small. 

\mypara{Qualitative evaluation.}
We further investigate how our full model handles two challenging scenarios.
The first one is when shortsighted behaviors lead to collisions, and the second one shows the robot freezing when it is close to both the goal and potential danger.

The scenario shown in Figure~\ref{fig:fig3a} is a typical example of the first scenario.
The one-step lookahead policy assigns almost equally high values to actions in the direction of $30^{\circ}$ and $210^{\circ}$, with $210^{\circ}$ being the highest.
This is reasonable for a shortsighted policy since no human occupies either of these two directions.
However, taking action around $210^{\circ}$ will result in a collision with human $\#$0.
By looking two steps ahead, the policy anticipates an unavoidable collision in two steps if it moves in the direction $210^{\circ}$ now.
The advantage of the multi-step lookahead policy also validates the effectiveness of encoding higher-order human interactions.
By observing the strong relation between human $\#$0 and human $\#$3, who are moving towards the same direction and are very close to each other, the robot predicts human $\#$3 will move to a direction of $15^{\circ}$ and predicts negative rewards for taking an action in the direction of $210^{\circ}$.
Thus, the two-step lookahead policy assigns low values to the direction of $210^{\circ}$ and avoids collisions by taking action in the direction of $30^{\circ}$.

The scenario shown in Figure~\ref{fig:fig3d} is a typical example of the second challenging scenario, where human $\#$3 is standing near the robot's goal.
The one-step lookahead policy assigns the highest value to the action in the direction of $225^{\circ}$ and a lower value to the action in the direction of $45^{\circ}$.
This is because taking an action towards $45^{\circ}$ will result in a discomfort penalty from stepping into the comfort zone of human $\#$3.
The two-step lookahead policy can predict the big positive reward after taking an action in the direction of $45^{\circ}$, and the positive reward two steps later can compensate the discomfort penalty in one step.
Thus, the two-step lookahead policy assigns the highest value to the action towards $45^{\circ}$ and makes a non-myopic decision.

\section{Real-world Demonstrations}

\begin{figure}
\centering
\includegraphics[width=0.2\textwidth]{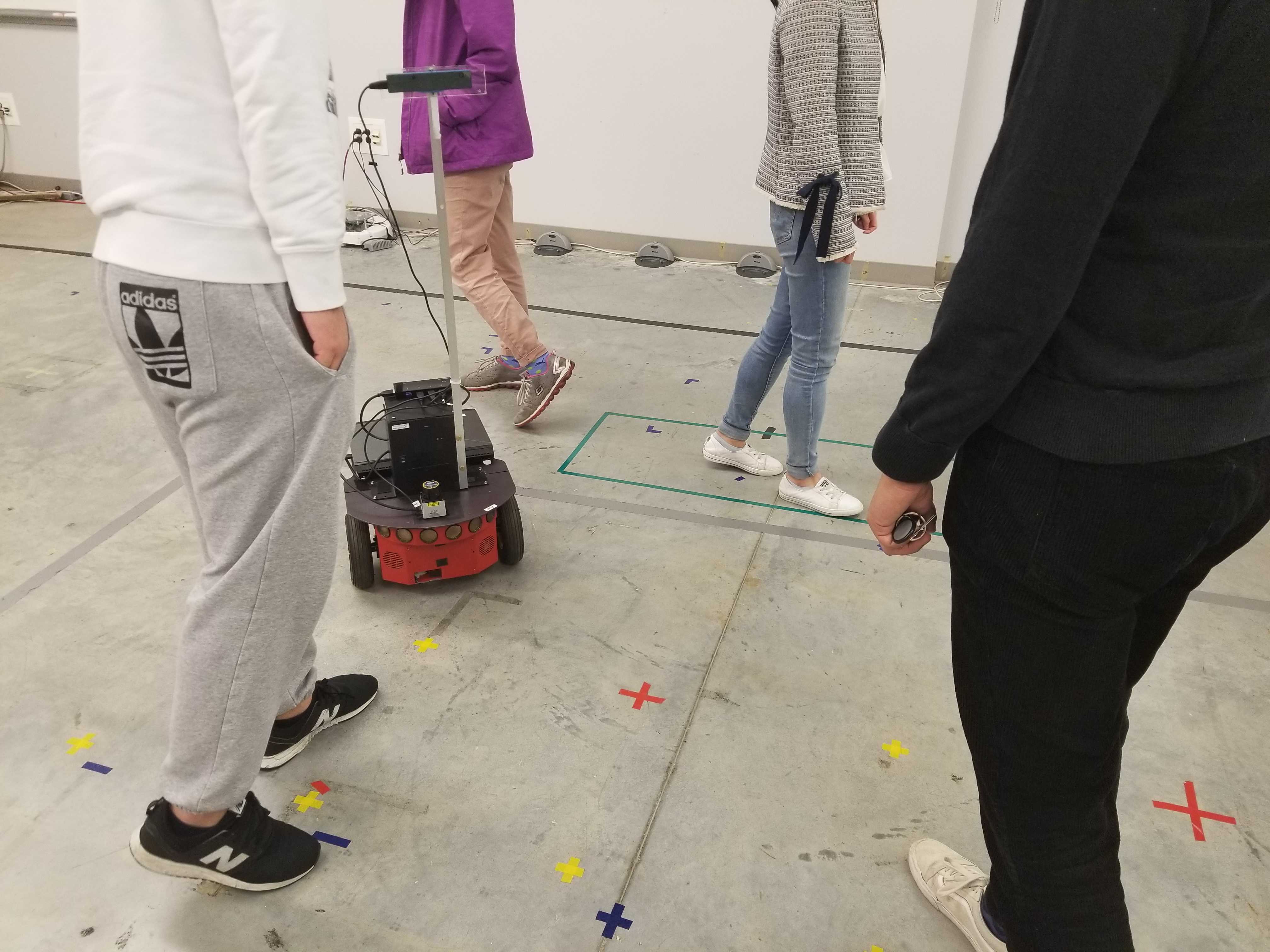} 
\caption{Pioneer robot used for real-world demonstrations.} \label{fig:robot}
\vspace{-0.7em}
\end{figure}

We deploy our trained policy on a Pioneer robotic platform equipped with an Intel RealSense ZR300 camera \shacamera{and a Hokuyo 2D LiDAR} (Figure \ref{fig:robot}). Test episodes are in the video. \shacamera{We first apply YOLO~\cite{redmon2016you} on depth data
to obtain 2D human positions. We then utilize an extended Kalman filter to track the human and compute human velocities.} This demonstration shows the potential of our method to model complex relations among interacting people in real scenes. Video: \url{https://youtu.be/U3quW30Eu3A}.

\section{Conclusion}
\label{sec:conclusion}
In this paper, we address crowd navigation with a relational graph learning approach.
By formulating human-robot and human-human interactions as a graph and using GCNs to compute interaction features, our model can estimate state values as well as predict human motion.
Augmented by $d$-step planning, our model explicitly plans into the future under a specific search budget.
We show our model outperforms baseline methods by a large margin and can handle challenging navigation scenarios.
The relational graph learning approach we proposed can be extended in several ways.
For example, we do not model the temporal dynamics of past human trajectories, which can help infer the intent of individuals and group structure among humans.

\small{\noindent\textbf{Acknowledgements} This research is partially supported by an NSERC Discovery Grant.}
\addtolength{\textheight}{-0cm}   

\def\url#1{}
\bibliographystyle{IEEEtran}
\bibliography{iros}

\end{document}